# BHRAM-IL: A Benchmark for Hallucination Recognition and Assessment in Multiple Indian Languages

**Hrishikesh Terdalkar** [†]
hrishikesh.rt@hyderabad.bits-pilani.ac.in

**Kirtan Bhojani** [‡]    **Aryan Dongare** [‡]    **Omm Aditya Behera** [‡]
{f20230366, f20230194, f20230434}@hyderabad.bits-pilani.ac.in

[†]Department of Computer Science and Information Systems
[‡]Department of Electrical and Elecronics Engineering
[†‡]BITS Pilani, Hyderabad Campus

## Abstract

Large language models (LLMs) are increasingly deployed in multilingual applications but often generate plausible yet incorrect or misleading outputs, known as hallucinations. While hallucination detection has been studied extensively in English, under-resourced Indian languages remain largely unexplored. We present **BHRAM-IL**, a **b**enchmark for **h**allucination **r**ecognition and **a**ssessment in **m**ultiple **I**ndian **l**anguages, covering Hindi, Gujarati, Marathi, Odia, along with English. The benchmark comprises 36,047 curated questions across nine categories spanning factual, numerical, reasoning, and linguistic tasks. We evaluate 14 state-of-the-art multilingual LLMs on a benchmark subset of 10,265 questions, analyzing cross-lingual and factual hallucinations across languages, models, scales, categories, and domains using category-specific metrics normalized to (0,1) range. Aggregation over all categories and models yields a primary score of 0.23 and a language-corrected fuzzy score of 0.385, demonstrating the usefulness of BHRAM-IL for hallucination-focused evaluation. The dataset, and the code for generation and evaluation are available on GitHub (https://github.com/sambhashana/BHRAM-IL/) and HuggingFace (https://huggingface.co/datasets/sambhashana/BHRAM-IL/) to support future research in multilingual hallucination detection and mitigation.

## 1 Introduction and Motivation

Large Language Models (LLMs) have rapidly become the backbone of modern NLP applications, excelling in tasks such as summarization, question answering, and conversational systems. However, they continue to suffer from a major limitation: the tendency to generate *hallucinations*—fluent but factually incorrect or misleading outputs. Hallucinations significantly undermine trust in LLMs, especially when they are deployed in sensitive, real-world applications.

Indian languages pose unique challenges due to rich morphology, diverse syntax, orthographic variations, and limited digital resources. Without proper benchmarks, the reliability of LLM outputs in these languages cannot be systematically assessed.

### 1.1 Scope and Contributions

In this paper, we introduce *BHRAM-IL*[1] a *multilingual evaluation benchmark* for hallucination recognition across four Indian languages and English. Our benchmark explicitly targets under-resourced languages, filling a critical gap in existing evaluation frameworks. We also analyze model performance, identify recurring patterns of hallucinations, and explore mitigation strategies tailored for these languages. Our contributions are: (1) a curated dataset of 36,047 questions across 9 categories, covering Hindi, Gujarati, Marathi, Odia, and English[2]. (2) a taxonomy of hallucination types: *Language Hallucination* (wrong language output) and *Factual Hallucination* (incorrect answers), and (3) thorough benchmarking of 14 state-of-the-art LLMs and varying prompt setups on these languages, including analysis of hallucination patterns across dimensions such as model, scale, language, category and domain.

## 2 Related Work

**Hallucination Studies in High-Resource Languages.** LLMs have been shown to generate hallucinations across diverse applications such as question answering, summarization, dialogue systems, and knowledge-grounded tasks. Several benchmark datasets, such as TruthfulQA (Evans et al., 2021), HaluEval (Li et al., 2023), and FActScore (Min et al., 2023), have been proposed to systematically measure factual consistency. Other works focus on task-specific hallucinations, e.g., XSumFaith (Jia et al., 2023) for summarization or WikiBio (Stranisci et al., 2023) for biographical generation. These efforts provide useful insights but remain limited to high-resource languages, particularly English, with some recent works extending to Chinese and European languages. A consistent theme across global studies is the lack of generalizable taxonomies for hallucination types across multilingual settings.

---

[1]The Sanskrit word भ्रम (*bhrama*) is approximately synonymous to *confusion* or *hallucination*.

[2]Union of the languages spoken by the authors.

**Indian Datasets.** In the multilingual context, studies such as X-FACT (Gupta and Srikumar, 2021) extend factuality evaluation to European and East Asian languages. For Indic languages, benchmarks like AI4Bharat datasets (Mhaske et al., 2022; Kakwani et al., 2020; Kunchukuttan et al., 2020), and PARIKSHA (Watts et al., 2024) provide multilingual evaluation suites, yet none explicitly targets hallucination tendencies. While they provide high-quality bilingual or monolingual corpora, they do not contain annotations or structures to evaluate factual consistency of model outputs, leaving a major gap in evaluating hallucinations in Indian languages. *BHRAM-IL* attempts to address this by combining and acquiring parallel multilingual data, conducting hallucination evaluation, and cross-prompt analysis for Indian languages.

**Limitations of Current Hallucination Datasets and Benchmarks.** Current hallucination benchmarks suffer from several limitations. A large majority are dominated by English and a handful of other high-resource languages, leaving low-resource Indic languages—such as Hindi, Gujarati, Marathi, and Odia—largely unaddressed. Most benchmarks are tied to a single task (e.g., summarization or open-domain QA), which makes it difficult to generalize hallucination findings across domains. Many existing datasets also rely on crowd-sourced judgments for factuality, which may lack consistency, especially in multilingual settings. To the best of our knowledge, no prior work provides a structured benchmark targeting hallucinations in Indian languages, despite the growing deployment of LLMs in Indian contexts.

## 3 Dataset

*BHRAM-IL* is a multilingual benchmark for hallucination analysis across five languages: Hindi (HI), Gujarati (GU), Marathi (MR), Odia (OR), along with English (EN). The benchmark targets two broad hallucination phenomena: (i) *language hallucination*: when a model produces an output in a language different from the input, and (ii) *factual hallucination*: when the output may be linguistically correct but factually incorrect.

LLMs exhibit different types of hallucinations depending on the task type. To capture this diversity, BHRAM-IL covers 9 task categories (§3.1) and a domain taxonomy that spans both global and India-specific knowledge areas (§A.1). Except for NER, all questions are *parallel* across five languages; NER is independently curated per language (§3.2.6).

### 3.1 Categories of Questions

The dataset includes questions designed to check the factual, numerical, reasoning, and linguistic abilities of LLMs. We choose a suitable primary evaluation metric for each category. Task descriptions, expected output formats, and metrics are summarized in Table 1.

#### 3.1.1 Factual

Hallucinations often manifest as incorrect information presented as fact, so factual questions are central to our benchmark. The `GenFact`, `IndFact`, and `T/F` categories contain factual questions drawn from approximately 30 domains such as geography, sports, literature etc. (Table 4 in §A.1).

#### 3.1.2 Numerical

LLMs, being "language models", do not possess the ability to perform numerical computation. Nevertheless, they have been shown to produce answers to mathematical questions. To measure this ability, we include numerical questions from seven fields of mathematics (Table 4 in §A.1).

#### 3.1.3 Reasoning

The ability to reason over factual knowledge is tested using chronological ordering (`Chrono`), multiple-choice reasoning questions (`Reasoning`), and semantically incorrect questions (`SemInc`). In `Chrono`, models are tasked with ordering historical events chronologically. `Reasoning` questions provide a scenario and a question with multiple possible answers, and models must choose the most appropriate one. `SemInc` contains semantically incorrect questions, e.g., "Who is the Prime Minister of Gujarat?", a grammatically correct yet semantically invalid question.[3]

### 3.1.4 Linguistic

We test linguistic abilities using Named Entity Recognition (`NER`) and Word Ordering (`WO`) tasks. Word Ordering questions present a sentence in a jumbled word order, and the model is asked to provide a correct word order.

## 3.2 Creation Pipelines

### 3.2.1 GenFact, IndFact and True/False (LLM-assisted from Wikipedia)

We assemble topic lists per domain (see §A.1). For each topic, we query Wikipedia and extract the first five paragraphs (introductory sections are typically high-precision). A controlled prompt asks an LLM to generate $n$ short, unambiguous questions per topic with single-span answers grounded in the provided text. We enforce templates that (i) avoid opinionated or ambiguous phrasing, (ii) prefer entity/date/quantity answers, and (iii) prohibit multi-hop or open-ended synthesis. *True/False* items are derived by flipping or preserving atomic facts from the same context (balanced sampling).

**Candidate filtering.** We discard questions that (a) lack a unique minimally sufficient answer in the context, (b) collapse to definition lookups likely to be ambiguous across languages, or (c) produce underspecified entities (e.g., missing disambiguating qualifiers).

### 3.2.2 Chrono (Rule-based from Wikipedia)

We harvest events (battles/wars) and canonical dates from Wikipedia/Wikidata. Events are rejected if any date is missing/ambiguous. Each item samples five distinct events; gold order is computed by sorting ISO-normalized dates. We preserve exact surface strings as options to avoid inadvertent hints in translation.

### 3.2.3 Maths (Curated)

We curate single-answer questions spanning *Algebra, Counting & Probability, Geometry, Intermediate Algebra, Number Theory, Pre-algebra, Precalculus* from (Awsaf, 2025). Gold answers are numeric or short symbolic forms. We standardize to ASCII numerals and permit benign formatting variants during evaluation (e.g., commas, trailing zeros).

### 3.2.4 Reasoning (Curated)

We select deductive/critical-reasoning items (MCQ/short answer) curated from (Liu et al., 2020). Each item has one correct option.

### 3.2.5 Semantically Incorrect (LLM-generated, Manual Curation)

One of the novel contributions of *BHRAM-IL* is the category of semantically incorrect questions. We use a high-capability LLM (GEMINI 2.5 PRO) (Comanici et al., 2025) to synthesize ill-posed prompts (category errors, anachronisms, geographically incongruous statements, false premises) with explicit constraints to avoid trivially nonsensical text (e.g., "How tall is sadness?"). We manually filter out repetitive cases, e.g. instances of 'Who is the prime minister of ___?' with different states.

### 3.2.6 NER (Curated, Non-Parallel)

The `NER` dataset was curated from (Mhaske et al., 2022). We compile NER sentences per Indic language from these existing resources with entity annotations. Sentences are *not parallel* across languages. We retain the original language's orthography and label schema. For our release, we include HI/GU/MR/OR (no EN).

### 3.2.7 Word Ordering (Curated, Parallel)

The parallel word/sentence ordering items were compiled by recognizing identical entries in the Hindi (HI), Gujarati (GU), Marathi (MR), Odia (OR), and English (EN) datasets from (Ramesh et al., 2022) and then aligning them across these languages. Each item has a canonical reference ordering per language, with alternative valid permutations retained when present in the source (rare; flagged in metadata).

## 3.3 Translation and Parallelization

All non-NER categories are translated from EN into HI/GU/MR/OR using a translation-only instruction to a high-capability LLM (GEMINI 2.5 PRO), explicitly disallowing transliteration and paraphrasing unless required by grammar. We enforce:

1. **Script adherence**: Devanagari (HI/MR), Gujarati (GU), Odia (OR); no Latinization except for proper nouns that are typically written in Latin script.

---

[3] In India, a state does not have a Prime Minister, but a Chief Minister.

| Category | Description | Output | Primary Metric |
|---|---|---|---|
| GenFact | Short factual questions various domains. | Short span (entity, number, phrase). | Exact Match |
| IndFact | India-centric factual questions. | Short span (entity, number, phrase). | Exact Match |
| T/F | Factual questions with binary answers. | True / False. | Exact Match |
| Chrono | Sort 5 events chronologically. | Comma-separated events. | Kendall's $\tau$. |
| Maths | Numerical/symbolic problem solving. | Numbers in English. | Exact Match |
| Reasoning | MCQ reasoning. | Correct option text. | Exact Match |
| SemInc | Detect ill-posed prompts. | 'Invalid' or factual span. | Exact Match |
| NER | Extract PER/LOC/ORG entities. | BIO tags. | F1 Score |
| WO | Reorder words into a sentence. | Coherent sentence. | Kendall's $\tau$ |

Table 1: Task definitions, outputs, and evaluation metrics

2. **Semantics fidelity**: preserve named entities, dates, and quantities; avoid introducing qualifiers not in the source.

3. **Answer consistency**: translated question must have the same gold answer (after language-specific rendering).

**Manual curation.** Due to the size of the dataset, bilingual annotators reviewed a stratified sample of approximately 10% of the items to assess translation fidelity and correctness. Their review helped identify common error patterns and informed minor automated cleaning steps. Comprehensive human verification of the entire dataset remains an important direction for future work, and we plan to incorporate full manual annotation in subsequent iterations.

### 3.4 Data Statistics

The resulting distribution is shown in Table 2. All categories except NER are parallelized across the five languages (HI/GU/MR/OR/EN). We currently benchmark roughly 25% of the collected questions due to resource constraints,[4] while releasing the full dataset for future evaluation. NER is non-parallel and covers the four Indian languages from this set (HI/GU/MR/OR). Figure 1 shows the distribution of data across languages.

## 4 Evaluation

Experiments were conducted on an H100 GPU machine (AMD EPYC 9354 host) and a macOS M2 Pro 15 laptop. Inference was performed using Ollama[5]. Larger models ($\geq$8B) were executed on the H100 GPU, while smaller or quantized variants were run on macOS M2 Pro hardware to enable broader coverage under limited resource and time constraints.

| Category | #Items (benchmark) | #Items (full) |
|---|---|---|
| GenFact | 1950 | 4870 |
| IndFact | 1135 | 5675 |
| T/F | 985 | 9825 |
| Chrono | 980 | 2450 |
| Maths | 875 | 875 |
| Reasoning | 705 | 705 |
| SemInc (Invalid) | 850 | 3620 |
| NER | 805 | 4017 |
| WO | 1005 | 4010 |
| **Total (core)** | **10,265** | **36,047** |

Table 2: Category-wise distribution. Counts denote total items after language replication; NER is non-parallel (sum over HI/GU/MR/OR); *#Items (benchmark)* set is used to establish the current benchmark.

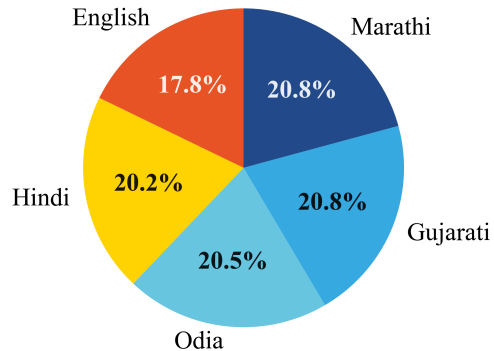

Figure 1: Distribution questions across languages.

### 4.1 Language Models

We evaluate a diverse set of open-weight LLMs spanning multiple parameter scales and architectures.

The GEMMA3 series[6] (270M, 1B, 4B, 12B, 27B) represents Google's latest family of instruction-tuned multimodal models with context lengths up to 128K tokens. These models serve as a scale-controlled baseline for multilingual robustness and hallucination sensitivity.

---
[4]Benchmarking on the entire dataset is ongoing.
[5]https://ollama.com
[6]https://ollama.com/library/gemma3

LLaMA 3.2 (3B)[7] and MISTRAL-NEMO (12B)[8] represent strong publicly available baselines for general-purpose reasoning and generation. Both are widely used in multilingual and factuality benchmarks; LLaMA 3.2 provides a balanced encoder-decoder alignment, while MISTRAL-NEMO offers optimized inference for efficiency-oriented deployment.

QWEN3 (8B)[9] is a multilingual foundation model trained on extensive cross-lingual corpora, including Indic languages, and has demonstrated competitive results (Yang et al., 2025).

Two Indic models, NAVARASA-2.0[10] and KRUTRIM-2[11], are included to assess performance on native-language data. Both are trained primarily on Indian languages; NAVARASA-2.0 (both FP16 and quantized Q4_K_M variants) emphasizes linguistic coverage across 11 Indic languages, whereas KRUTRIM-2 (FP16 and Q4_K_M) targets factual accuracy and instruction following in bilingual (EN-Indic) settings. Both models were consistently among top-10 performers in PARIKSHA (Watts et al., 2024) benchmark.

Finally, GPT-OSS (20B) and GPT-OSS (120B)[12] are open-weight reasoning models that employ a Mixture-of-Experts (MoE) architecture with MXFP4 quantization, offering competitive performance on reasoning and multilingual understanding tasks.

## 4.2 Prompting Strategies

Prompt design is critical in multilingual LLM evaluation. We compare two prompting strategies:

- **English prompts:** the instruction and task description are in English, while the question may be in any of the five target languages.

- **Native prompts:** the instruction and description are in the same language as the question (Hindi for Hindi questions, Marathi for Marathi, etc.).

We report hallucination rates and accuracy under both strategies, isolating how prompt language influences model stability and error modes.

---

[7] https://ollama.com/library/llama3.2
[8] https://ollama.com/library/mistral-nemo
[9] https://ollama.com/library/qwen3
[10] hhttps://huggingface.co/collections/Telugu-LLM-Labs/navarasa-20-models
[11] https://huggingface.co/krutrim-ai-labs
[12] https://ollama.com/library/gpt-oss

### 4.2.1 Prompting Text Completion Models

Some evaluated models (NAVARASA-2.0 and variants) are pure text completion models rather than chat-style models. These models often produced empty or malformed outputs when given the same prompt structure we used for chat models. To mitigate this, we reformatted prompts (§B) to coax them into producing valid, structured responses. This heuristic adaptation improved yield and allowed us to include them in the evaluation.

## 4.3 Hallucination Types and Classification

We distinguish two primary hallucination classes:

- **Language hallucination:** occurs when the model responds in a language different from the input prompt, despite a system instruction to output in the same language. We flag any such mis-language response (commonly defaulting to English) as language hallucination.

- **Factual hallucination:** occurs when the output is in the correct language but is factually incorrect relative to the gold reference.

We embed a system prompt instructing each model to respond in the same language as the question. Violations of that instruction are recorded as language hallucination. Outputs that remain in the correct language but deviate from the ground truth are recorded as factual hallucination.

## 4.4 Evaluation Metrics

We evaluate each category using task-appropriate metrics, as described in §3.1: Exact Match (EM) for span-based factual tasks, F1 for extraction (NER), and Kendall's $\tau$ for ordering tasks. We treat these as the *primary scores* (PS). For each prediction, we also record whether the model answered in the designated language. If it responds in a different language, we realign the output to the corresponding gold answer in that language (when available) and recompute the primary metric to obtain a *language-corrected score* (LCS). In both settings, Fuzzy Match uses normalized string similarity with a fixed threshold to allow minor lexical variation. All metrics are computed after normalization (Unicode NFC, whitespace and punctuation trimming), as defined in §3.3.

## 5 Results

We now present model performance across hallucination metrics and task categories.

| Model | English Prompts | | | Native Prompts | | |
|---|---|---|---|---|---|---|
| | LH% | PS | LCFS | LH% | PS | LCFS |
| LLaMa3.2:3b | 24.54 | 0.16 | 0.33 | 16.16 | 0.13 | 0.27 |
| Qwen3:8b | 29.11 | 0.42 | 0.60 | 21.85 | 0.35 | 0.56 |
| Mistral-NeMo:12b | 36.05 | 0.19 | 0.35 | 41.79 | 0.11 | 0.25 |
| Gemma3:270m | 47.20 | 0.13 | 0.26 | 20.38 | 0.09 | 0.18 |
| Gemma3:1b | 42.11 | 0.17 | 0.31 | 43.99 | 0.13 | 0.22 |
| Gemma3:4b | 21.74 | 0.23 | 0.40 | 18.75 | 0.17 | 0.34 |
| Gemma3:12b | 22.42 | 0.31 | 0.51 | 18.37 | 0.27 | 0.45 |
| Gemma3:27b | 23.04 | 0.41 | 0.58 | 18.28 | 0.37 | 0.55 |
| Navarasa2.0:Q4_K_M | 28.77 | 0.07 | 0.20 | 20.34 | 0.06 | 0.16 |
| Navarasa2.0:FP16 | 31.35 | 0.07 | 0.20 | 20.90 | 0.06 | 0.17 |
| Krutrim2:Q4_K_M | 23.97 | 0.30 | 0.49 | 17.13 | 0.27 | 0.46 |
| Krutrim2:F16 | 28.21 | 0.29 | 0.49 | 17.92 | 0.29 | 0.48 |
| GPT-OSS:20b | 25.74 | 0.40 | 0.55 | 27.16 | 0.36 | 0.53 |
| GPT-OSS:120b | 28.92 | 0.44 | 0.61 | 28.58 | 0.40 | 0.58 |

Table 3: Overall performance aggregated across all categories per model. (LH%: Language Hallucination %, PS: Primary Score, LCFS: Language Corrected Fuzzy Score)

## 5.1 Overall Hallucination Rates

Table 3 reports, for each model and prompting strategy, the rates of language hallucination, and factual hallucination measured using primary score and language-corrected fuzzy score metrics.

Native prompting consistently reduces language hallucination rates across most models, with two notable exceptions: GEMMA3:1B and MISTRAL-NEMO:12B. Smaller GEMMA3 variants (270M, 1B) exhibit particularly high language hallucination when prompted in English, frequently defaulting to English responses (Figure 4). In contrast, GPT-OSS models maintain relatively stable performance across both prompting styles.

The top-performing models are the GPT-OSS series, QWEN3, and the larger GEMMA3 variants (12B, 27B), followed by KRUTRIM2. Among these, QWEN3 demonstrates exceptional parameter efficiency. Notably, even the best primary score remains low at 0.44, with the language-corrected fuzzy score reaching only 0.61, reflecting persistent hallucination challenges and validating the benchmark's utility.

## 5.2 Language vs Category

Figure 2 visualizes task-appropriate metrics per category and language as a heatmap. For GenFact, most models obtain relatively high scores with only minor errors on rare entities. IndFact is slightly harder: models often mis-handle orthography, inconsistently translate names, or hallucinate local facts. Chrono exhibits strong variation across languages, with English questions performing the best and Marathi performing the worst. Maths performance is fairly uniform (around 0.23) across languages and prompt types. Reasoning MCQs generally achieve high accuracy; native prompts slightly improve scores for most languages, but Odia drops from 0.41 to 0.34. SemInc shows a marked gap between English (0.57) and the Indian languages (around 0.38), and Odia again degrades under native prompting (0.37 to 0.27). NER accuracy varies widely, with the best performance in Odia (0.55) and the lowest in Hindi (0.36). WO also shows cross-lingual variation (0.34–0.47), with a modest decrease when using native prompts (0.26–0.43). Overall, T/F questions prove to be the easiest category, achieving the highest scores across languages.

## 5.3 Model vs Category

Category-wise performance across models is depicted in Figure 7 in §C.1. GPT-OSS models demonstrate superior performance in most categories, with QWEN3 and the larger GEMMA3 variants (12B and 27B) following closely. Maths category exhibits the most pronounced performance gap: leading models (GPT-OSS and QWEN3) achieve scores exceeding 0.8, while other models fall below 0.3. Notably, the 8B parameter QWEN3 model (0.83) outperforms the 20B parameter GPT-OSS model (0.80) and nearly matches the 120B GPT-OSS variant, highlighting exceptional parameter efficiency. Chronological ordering emerges as the most challenging category, with even the top-performing QWEN3 model scoring below 0.4. Other difficult categories include `IndFact`, `WO`, and `Maths`. The NAVARASA model series consistently underperforms across all categories.

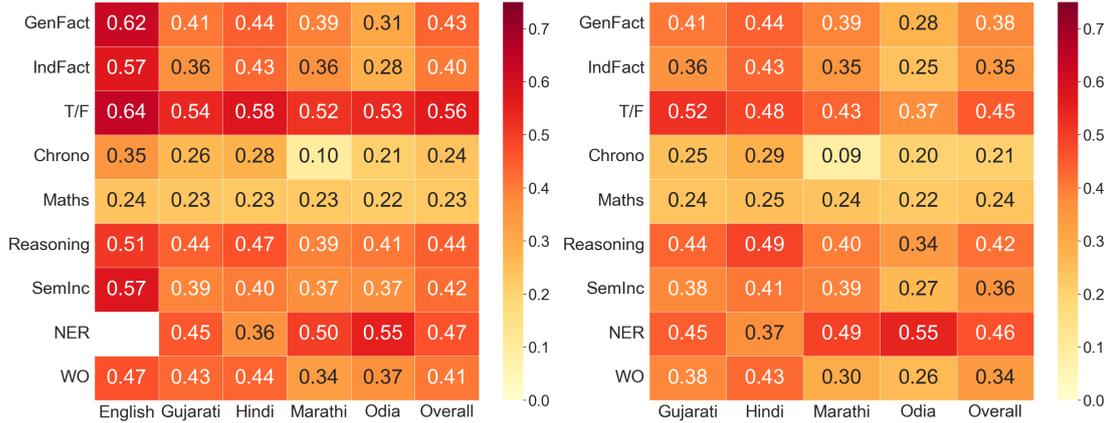

Figure 2: Cumulative performance of models by language and category with English (left) and native (right) prompts based on averaged language-corrected fuzzy score.

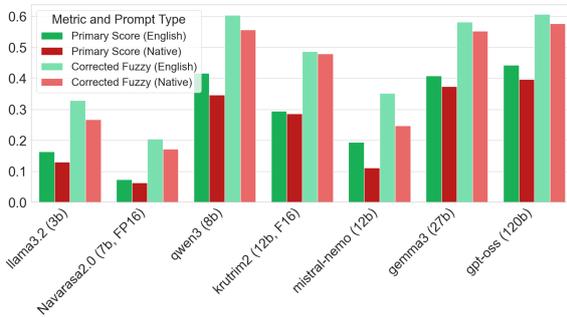

Figure 3: Comparison of the largest benchmarked models in each series of mdoels

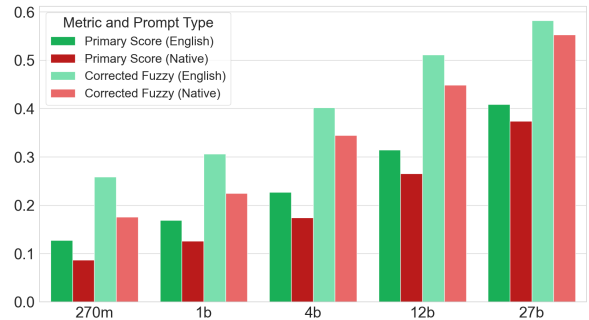

Figure 4: Comparison of GEMMA3 models by number of parameters.

### 5.4 Model vs Language

Analysis reveals that models achieve their highest performance on English questions. GPT-OSS, larger GEMMA3 variants, and QWEN3 demonstrate the strongest multilingual capabilities. While trailing the top performers, KRUTRIM2 consistently outperforms other models across languages, showing relatively uniform performance. Hindi and Gujarati exhibit marginally better results than Marathi and Odia. Figures 8 and 9 in §C.2 illustrates these trends.

### 5.5 Effect of Prompt Language

We also investigate the effect of English versus native prompts (Table 3, Figure 2). English questions yield the strongest performance overall; among the Indian languages, Hindi performs best, while Odia shows a consistent drop across categories under native prompting. Native prompts yield slight accuracy improvements for Hindi and Marathi, but marginal decreases for Gujarati and Odia. Native prompts reduce language hallucinations, especially for smaller models (Table 3). However, they also lead to a drop in performance compared to English prompts (Figure 3). As shown in Figure 5 in §C, this accuracy drop is smaller for Indic models such as NAVARASA-2.0 and particularly KRUTRIM2 than for non-Indic models with a similar number of parameters.

### 5.6 Summary of Findings

Our evaluation reveals several key findings. Native prompting substantially reduces language hallucination rates across most models, establishing clear performance tiers: GPT-OSS, QWEN3, and larger GEMMA3 variants lead in multilingual capability, yet even these top performers achieve modest scores (primary: 0.44, corrected: 0.61), indicating persistent hallucination challenges. Performance follows a linguistic hierarchy with En-glish outperforming Indian languages, Hindi leading among Indic languages, and Odia showing the greatest degradation. Category-wise, chronological ordering proves most difficult, while mathematics exhibits the widest performance gap between leading and trailing models. Notably, QWEN3 demonstrates exceptional parameter efficiency, and KRUTRIM2 maintains consistent cross-language performance despite not leading in absolute scores. These results highlight the need for improved multilingual alignment and suggest directions for model design, prompt strategy, and future benchmarks through dataset expansion, and retrieval-augmented approaches.

## 6 Discussion

Overall, the results reveal complementary strengths and weaknesses across models and task types rather than a single dominant frontier model.

**Behaviour in SemInc category.** We observe a pronounced drop in accuracy for the 'False Premise' subcategory of `SemInc`, especially in larger models (e.g., GPT-OSS 20B). For example, questions such as '*As the Earth is flat, what is at the edge of the Earth?*' require the model to reject the presupposition rather than answer it literally. The sharp performance drop suggests that model scale alone does not guarantee robustness to semantically inconsistent prompts: larger models often prioritize coherent continuation of the premise over challenging it. Addressing this weakness likely requires targeted training signals, explicit mechanisms for contradiction handling, and evaluation datasets that encode subtler forms of semantic inconsistency.

**Contrasting performance on WO and Maths categories.** GPT-OSS models achieve high accuracy in `Maths` outperforming other models, but perform poorly on word ordering, whereas most other models outperform the GPT-OSS models in `WO`. This contrast highlights architectural and training differences that lead to domain-specific strengths. Our results suggest that pretraining data and optimization strategies shape distinct reasoning biases, and future work should investigate how to combine these capabilities within a single model.

**Indic vs general-purpose models.** For `GenFact` questions, GPT-OSS and GEMMA3 models outperform KRUTRIM2, but on `IndFact` questions KRUTRIM2 performs slightly better, indicating that Indic-focused models better capture localized knowledge. This pattern underscores the importance of regionally diverse training data for evaluating and deploying LLMs in Indian contexts.

## 7 Conclusion and Future Work

We introduced **BHRAM-IL**, the first large-scale multilingual benchmark for hallucination detection in Indian languages. The dataset spans nine task categories and five languages—Hindi, Gujarati, Marathi, Odia, along with English—covering both *language* and *factual* hallucination phenomena. Through systematic evaluation of 14 language models, ranging from compact (270M) to large-scale (120B) architectures, we observed clear dependencies between model size, multilingual training coverage, and hallucination behaviour. Larger multilingual models such as GEMMA3 27B, GPT-OSS 120B, and QWEN3 8B achieve higher factual accuracy and lower language drift, whereas Indic-centric models like KRUTRIM2 and NAVARASA-2.0 maintain strong script fidelity but weaker factual grounding. Our results show that using native-language prompts leads to an overall drop in accuracy across models. However, this decline is substantially smaller for Indian LLMs compared to others, indicating that these models are better aligned with native-language inputs. This highlights the need for stronger multilingual alignment in non-Indian models. The dataset, code, and evaluation results, are released via *GitHub* (https://github.com/sambhashana/BHRAM-IL/) and *HuggingFace* (https://huggingface.co/datasets/sambhashana/BHRAM-IL/) to facilitate reproducibility and community benchmarking.

### 7.1 Future Directions

BHRAM-IL benchmark can serve as (i) a diagnostic suite for multilingual hallucination analysis, (ii) a resource for training hallucination detectors or reward models, and (iii) a foundation for cross-lingual alignment and trustworthiness studies. We plan to expand coverage to additional Indian languages (e.g., Tamil, Bengali, Telugu) and domains such as summarization, translation, and dialogue grounding. Future iterations will also incorporate automatic hallucination annotation models and human-in-the-loop verification to refine scoring and linguistic fidelity.

## Acknowledgements

We thank the anonymous reviewers for their constructive feedback and valuable suggestions. This work was supported by the High Performance Computing (HPC) facilities at BITS Pilani, Hyderabad Campus. We also acknowledge the ACL BHASHA workshop organizers for providing a platform for research in Indian language NLP.

## Limitations

Although **BHRAM-IL** provides the first systematic framework for evaluating hallucinations in Indian languages, several limitations remain.

First, most non-English data rely on machine translation with partial manual review; subtle semantic drift or culturally biased renderings may persist, and full human verification is still pending.

Second, coverage is limited to five languages and nine task categories, and the current benchmark uses only a subset of the collected data; other low-resource languages (e.g., Bengali, Tamil, Telugu, Kannada, Assamese, etc.) and additional domains remain out of scope.

Third, automatic metrics (Exact Match, Fuzzy Match, Kendall's $\tau$, F1) may miss pragmatic appropriateness, partial credit, or reasoning failures, and we have not yet included human evaluations of hallucination severity.

Fourth, inference was run on mixed hardware (H100 GPU and macOS M2 Pro) with quantized variants, which can introduce variability in generation quality and hallucination patterns.

Fifth, prompt design was fixed at inference time; we did not sweep decoding parameters, retrieval augmentation, or adversarial prompting, so robustness under alternative setups is untested. We view these gaps as priorities for the next release of the benchmark.

## A  Dataset

### A.1  Domain Taxonomy

To enable granular analysis, we categorize questions from `GenFact`, `IndFact`, `SemInc`, `Reasoning`, and `Maths` into specific domains, as detailed in Table 4. This classification supports targeted evaluation of model performance across knowledge areas.

We maintain domain balance by capping items per topic and ensuring representation across entity-centric and numeric/date-centric questions.

### A.2  Reproducibility Framework

To facilitate replication and extension, we provide:

- **Prompt templates** for both data generation and model evaluation in all languages.
- **Text normalization utilities** handling Unicode NFC, punctuation standardization, and Indic numeral conversion.
- **Evaluation scripts** implementing the metrics defined in § 3.1.

### A.3  Release Format

The dataset is released in JSONL format with the following schema:

- `question_id`: an identifier for questions shared across the five languages for parallel items.
- `language`: ISO 639-1 standard codes for the corresponding language (one of `en`, `hi`, `gu`, `mr`, `or`).
- `category`: category of the question (one of `factual_questions`, `indian_questions`, `true_false_questions`, `ner_questions`, `chrono_questions`, `maths_questions`, `semantically_incorrect_questions`, `word_ordering_questions`, `reasoning_questions`).
- `domain`: domain label for relevant categories (see §A.1).
- `question`: task-specific input text (question, options, sentence).
- `expected`: the ground truth answer (span/label/order), in the same language as input.

## B  Prompt Design Examples

We design language-specific prompts for each category, providing output format specifications to ensure structured responses. Complete prompt sets are available in the repository[13]. Below we showcase the chronological ordering (`Chrono`) prompts across all languages.

**English**
```
Order the following events chrono-
logically. Your response should only
contain the events as named in the
question itself, separated by commas.

Question: {question}
Output Format: {output_format}
```

**Hindi**
निम्नलिखित घटनाओं को कालानुक्रमिक रूप से व्यवस्थित करें। आपकी प्रतिक्रिया में केवल प्रश्न में नामित घटनाएं होनी चाहिए, जो अल्पविराम से अलग की गई हों:

प्रश्न: {question}
आउटपुट प्रारूप: {output_format}

**Gujarati**
નીચેની ઘટનાઓને કાળક્રમ મુજબ ગોઠવો. તમારા પ્રતિભાવમાં ફક્ત પ્રશ્નમાં નામિત ઘટનાઓ જ હોવી જોઈએ, જે અલ્પવિરામ દ્વારા અલગ કરવામાં આવી હોય:

પ્રશ્ન: {question}
આઉટપુટ ફોર્મેટ: {output_format}

---
[13] https://github.com/sambhashana/BHRAM-IL/

| GenFact Domains | | | |
|---|---|---|---|
| Art & Architecture | Festivals & Culture | Indian Classical Music | Physics |
| Cinema | World Geography | Inventions & Discoveries | Chemistry |
| Economics & Business | Literature | Space & Astronomy | Biology |
| Environment & Climate Change | Medicine & Health | Sports | Mathematics |
| Famous Personalities | Technology & Internet | World History | |
| IndFact Domains | | | |
| Indian History | Indian Geography (Political) | Indian Geography (Physical) | Indian Economy and Business |
| Indian Culture and Arts | Indian Sports | Mythology and Religions | Science & Technology in India |
| Indian Constitution and Politics | Indian Social Structures & Reform Movements | | |
| Reasoning Subcategories | | | |
| Critical Thinking | Logical Reasoning | Quantitative Reasoning | Scientific Reasoning |
| Verbal Reasoning | | | |
| SemInc Subcategories | | | |
| Invalid Role-Entity Pairing | Anachronistic | Geographically Incongruous | False Premise |
| Maths Subcategories | | | |
| Algebra | Geometry | Number Theory | Prealgebra |
| Counting & Probability | Intermediate Algebra | Precalculus | |

Table 4: Domains and Subcategories Across All Dataset Types

**Marathi**

खालील घटना कालक्रमानुसार लावा. तुमच्या प्रतिसादात फक्त प्रश्नात नमूद केलेल्या घटना असाव्यात, ज्या स्वल्पविरामाने वेगळ्या केल्या असतील:

प्रश्न: {question}
आउटपुट स्वरूप: {output_format}

**Odia**

ନିମ୍ନଲିଖିତ ଘଟଣାଗୁଡ଼ିକୁ କାଳାନୁକ୍ରମିକ ଭାବେ ସଜାନ୍ତୁ। ଆପଣଙ୍କ ପ୍ରତିକ୍ରିୟାରେ କେବଳ ପ୍ରଶ୍ନରେ ଦିଆଯାଇଥିବା ଘଟଣାଗୁଡ଼ିକ ରହିବା ଉଚିତ୍, ଯାହା କମା ଦ୍ୱାରା ପୃଥକ ହୋଇଥିବ:

ପ୍ରଶ୍ନ: {question}
ଆଉଟପୁଟ୍ ଫର୍ମାଟ୍: {output_format}

For text completion models (e.g., Navarasa-2.0), the following structured prompt format was used:

```
### Instruction:
{system_prompt}

Response Format:
{output_format}

{user_prompt_template}

### Input:
{question}

### Response:
```

## C  Comprehensive Results Analysis

Figure 5 showcases the performance of all models across both prompting strategies, showing primary and language-corrected fuzzy scores averaged over all categories. Figure 6 presents aggregate performance by category averaged over all models.

### C.1  Model-Category Interactions

Figure 7 compares performance of all 14 models across the 9 categories for English language prompts, revealing that scaling improves performance on factual categories (T/F, SemInc, Gen-Fact, IndFact) but provides diminishing returns for reasoning-intensive tasks (Maths, Reasoning, WO, Chrono). The results indicates that these tasks require deeper algorithmic reasoning rather than scale-driven pattern learning. Notably, QWEN3:8B achieves Maths performance (0.84) comparable to larger models like GPT-OSS:20B (0.79), and GPT-OSS:120B (0.84) demonstrating exceptional parameter efficiency.

### C.2  Cross-Lingual Performance Patterns

Figures 8 and 9 show consistent performance degradation from English to Indian languages. MISTRAL-NEMO:12B exhibits the steepest cross-lingual drop (0.58 in English vs 0.26–0.33 in Indian languages), while GPT-OSS:120B and GEMMA3:27B maintain stronger multilingual consistency.

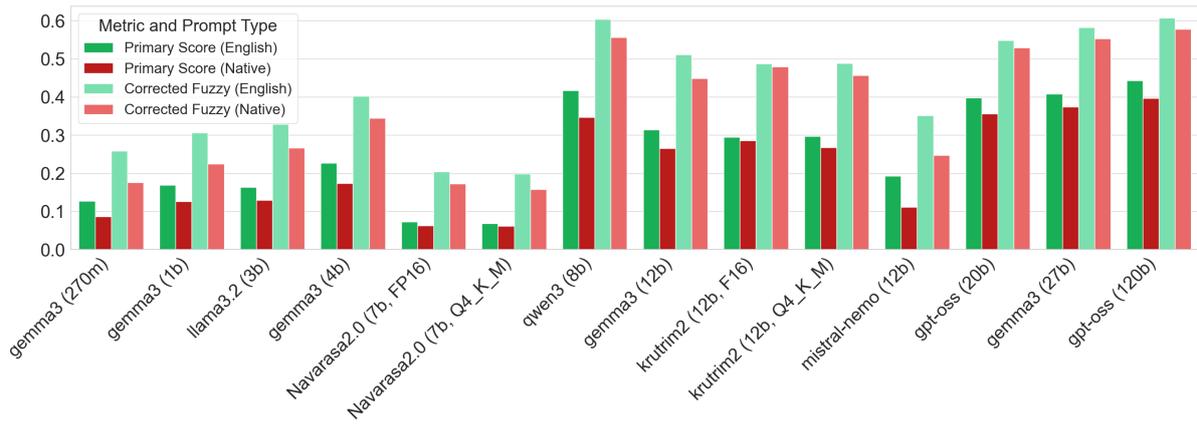

Figure 5: Overall performance of all models

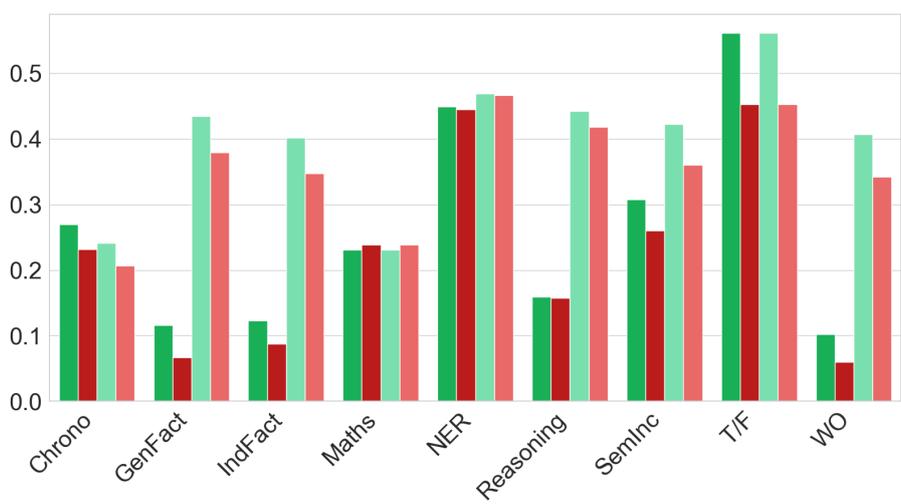

Figure 6: Overall Scores by Category ■ PS (English), ■ LCFS (English), ■ PS (Native), ■ LCFS (Native).

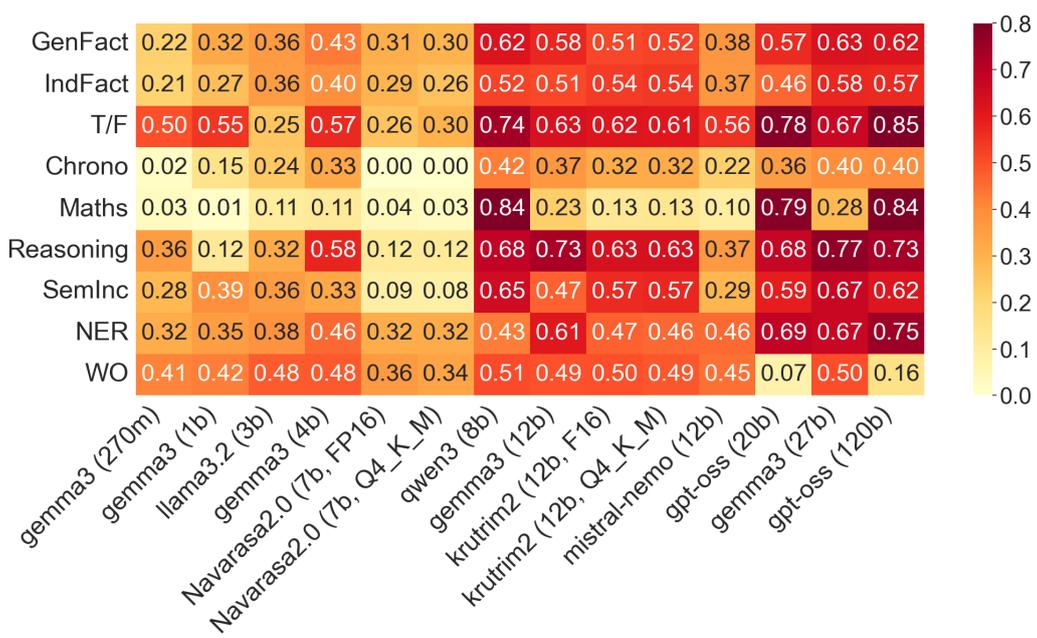

Figure 7: Performance of models across categories for English prompts

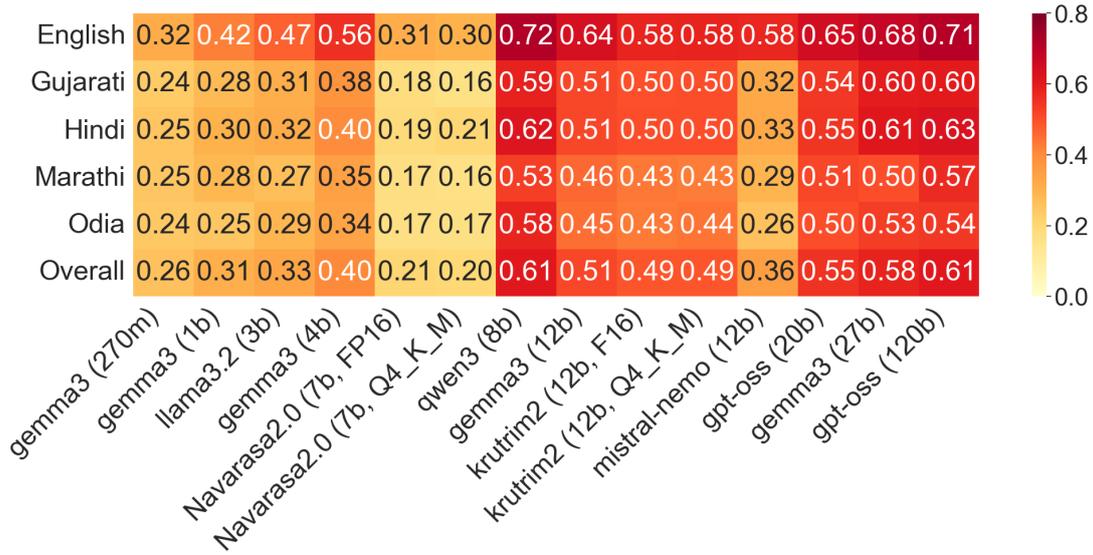

Figure 8: Performance of models across languages for English prompts

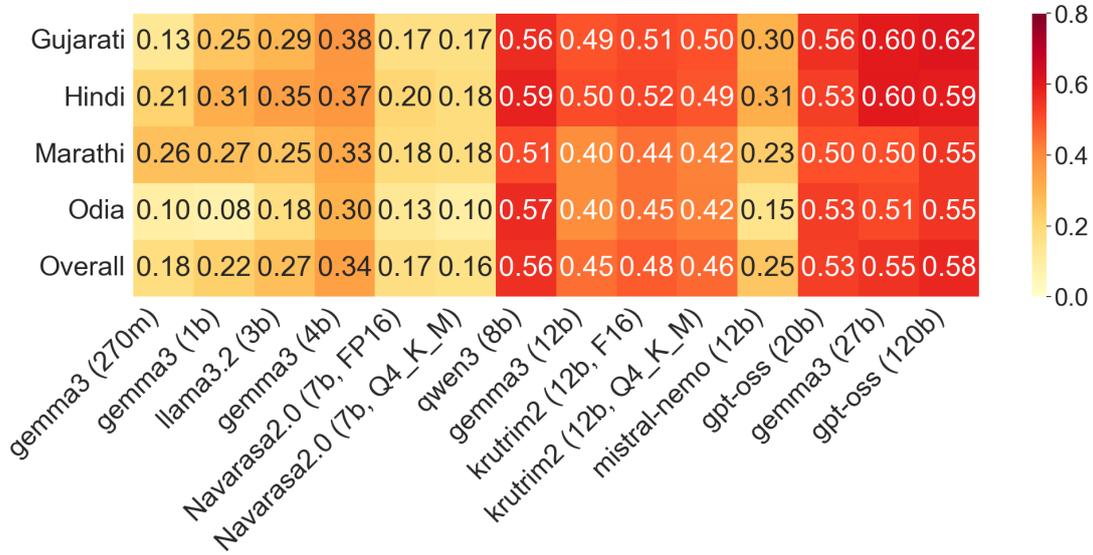

Figure 9: Performance of models across languages for Native prompts

### C.3 Domain-wise Performance Breakdown

Detailed domain-wise performance metrics (Tables 5–10) reveal systematic variations in hallucination rates and accuracy across certain knowledge domains.

The 'Technology & Internet' domain exhibits particularly high language hallucination rates, yet maintains strong corrected fuzzy scores. This suggests models struggle with language fidelity in technical domains while retaining factual knowledge.

Comparative analysis of GenFact and corresponding SemInc questions reveals performance degradation, indicating that semantic perturbations cause models to misclassify valid questions as 'Invalid;. This sensitivity to prompt framing highlights model brittleness in handling nuanced semantic variations.

Within the SemInc category, 'False Premise' questions show the most substantial performance decline. Models frequently accept assertively stated false information as true, demonstrating vulnerability to presupposition errors.

The Reasoning category reveals significant disparities, with quantitative reasoning substantially underperforming other subcategories. This indicates particular challenges in numerical and mathematical reasoning compared to verbal or logical reasoning tasks.

| Domain | Language Hallucination % | Primary Score | Corrected Fuzzy Score |
| --- | --- | --- | --- |
| Art & Architecture | 38.59 | 0.07 | 0.36 |
| Biology | 36.15 | 0.09 | 0.41 |
| Chemistry | 38.89 | 0.12 | 0.47 |
| Cinema | 36.48 | 0.06 | 0.35 |
| Economics & Business | 33.70 | 0.08 | 0.39 |
| Environment & Climate Change | 38.96 | 0.14 | 0.47 |
| Famous Personalities | 36.76 | 0.10 | 0.42 |
| Festivals & Culture | 35.80 | 0.09 | 0.38 |
| Indian Classical Music | 32.70 | 0.06 | 0.34 |
| Inventions & Discoveries | 35.39 | 0.10 | 0.43 |
| Literature | 33.14 | 0.11 | 0.37 |
| Mathematics | 38.64 | 0.09 | 0.41 |
| Medicine & Health | 35.04 | 0.07 | 0.37 |
| Physics | 33.46 | 0.13 | 0.44 |
| Space & Astronomy | 37.43 | 0.10 | 0.42 |
| Sports | 37.32 | 0.10 | 0.42 |
| Technology & Internet | 46.27 | 0.09 | 0.51 |
| World Geography | 34.67 | 0.07 | 0.38 |
| World History | 37.64 | 0.10 | 0.45 |

Table 5: Overall aggregated domain-wise performance for `GenFact`

| Domain | Language Hallucination % | Primary Score | Corrected Fuzzy Score |
| --- | --- | --- | --- |
| Indian Constitution and Politics | 31.14 | 0.11 | 0.36 |
| Indian Culture and Arts | 31.70 | 0.11 | 0.37 |
| Indian Economy and Business | 30.52 | 0.14 | 0.42 |
| Indian Geography (Physical) | 28.23 | 0.10 | 0.38 |
| Indian Geography (Political) | 31.77 | 0.09 | 0.36 |
| Indian History | 33.24 | 0.13 | 0.41 |
| Indian Mythology and Religions | 33.96 | 0.13 | 0.42 |
| Indian Social Structures & Reform Movements | 27.75 | 0.07 | 0.32 |
| Indian Sports | 33.07 | 0.11 | 0.38 |
| Science & Technology in India | 29.35 | 0.09 | 0.34 |

Table 6: Overall aggregated domain-wise performance for `IndFact`

| Domain | Language Hallucination % | Primary Score | Corrected Fuzzy Score |
| --- | --- | --- | --- |
| Art & Architecture | 53.71 | 0.48 | 0.48 |
| Biology | 52.79 | 0.46 | 0.46 |
| Chemistry | 53.15 | 0.46 | 0.46 |
| Cinema | 53.71 | 0.54 | 0.54 |
| Economics & Business | 53.57 | 0.55 | 0.55 |
| Environment & Climate Change | 53.64 | 0.45 | 0.45 |
| Famous Personalities | 52.79 | 0.51 | 0.51 |
| Festivals & Culture | 53.10 | 0.54 | 0.54 |
| Indian Classical Music | 53.14 | 0.52 | 0.52 |
| Inventions & Discoveries | 54.07 | 0.53 | 0.53 |
| Literature | 52.57 | 0.47 | 0.47 |
| Mathematics | 54.07 | 0.52 | 0.52 |
| Medicine & Health | 54.14 | 0.56 | 0.56 |
| Physics | 53.12 | 0.54 | 0.54 |
| Space & Astronomy | 52.57 | 0.47 | 0.47 |
| Sports | 53.02 | 0.51 | 0.51 |
| Technology & Internet | 53.57 | 0.58 | 0.58 |
| World Geography | 52.14 | 0.52 | 0.52 |
| World History | 53.29 | 0.59 | 0.59 |

Table 7: Overall aggregated domain-wise performance for `T/F`

| Domain | Language Hallucination % | Primary Score | Corrected Fuzzy Score |
| --- | --- | --- | --- |
| Algebra | 19.40 | 0.24 | 0.24 |
| Counting & Probability | 11.49 | 0.22 | 0.22 |
| Geometry | 15.71 | 0.23 | 0.23 |
| Intermediate Algebra | 22.71 | 0.20 | 0.20 |
| Number Theory | 11.89 | 0.23 | 0.23 |
| Prealgebra | 14.29 | 0.25 | 0.25 |
| Precalculus | 21.09 | 0.26 | 0.26 |

Table 8: Overall aggregated domain-wise performance for `Maths`

| Domain | Language Hallucination % | Primary Score | Corrected Fuzzy Score |
| --- | --- | --- | --- |
| Critical Thinking | 13.69 | 0.19 | 0.48 |
| Logical Reasoning | 14.83 | 0.13 | 0.41 |
| Quantitative Reasoning | 13.00 | 0.05 | 0.33 |
| Scientific Reasoning | 10.54 | 0.13 | 0.42 |
| Verbal Reasoning | 14.62 | 0.16 | 0.41 |

Table 9: Overall aggregated domain-wise performance for `Reasoning`

| Domain | Language Hallucination % | Primary Score | Corrected Fuzzy Score |
| --- | --- | --- | --- |
| Anachronistic | 12.06 | 0.60 | 0.60 |
| False Premise | 13.91 | 0.40 | 0.40 |
| Geographically Incongruous | 13.93 | 0.51 | 0.51 |
| Invalid Role-Entity Pairing | 16.74 | 0.51 | 0.51 |
| Art & Architecture | 25.45 | 0.08 | 0.27 |
| Biology | 28.33 | 0.11 | 0.32 |
| Chemistry | 27.69 | 0.09 | 0.26 |
| Cinema | 31.14 | 0.07 | 0.29 |
| Economics & Business | 31.44 | 0.07 | 0.32 |
| Environment & Climate Change | 24.02 | 0.10 | 0.28 |
| Famous Personalities | 24.51 | 0.13 | 0.32 |
| Festivals & Culture | 24.41 | 0.10 | 0.30 |
| Indian Classical Music | 24.83 | 0.04 | 0.23 |
| Inventions & Discoveries | 26.97 | 0.04 | 0.21 |
| Literature | 26.74 | 0.11 | 0.34 |
| Mathematics | 28.03 | 0.06 | 0.26 |
| Medicine & Health | 27.73 | 0.09 | 0.30 |
| Physics | 30.23 | 0.12 | 0.37 |
| Space & Astronomy | 21.29 | 0.06 | 0.26 |
| Sports | 31.89 | 0.08 | 0.33 |
| Technology & Internet | 22.73 | 0.11 | 0.35 |
| World Geography | 26.82 | 0.06 | 0.26 |
| World History | 26.06 | 0.08 | 0.30 |

Table 10: Overall aggregated domain-wise performance for `SemInc`